\documentclass[journal,twoside,web]{ieeecolor}
\usepackage{generic}
\usepackage{cite}
\usepackage{amsmath,amssymb,amsfonts}
\usepackage{algorithmic}
\usepackage{graphicx}
\usepackage{algorithm,algorithmic}
\usepackage{hyperref}
\usepackage{graphicx}
\usepackage{amsmath}
\usepackage{amssymb}
\usepackage{dsfont}
\usepackage{float}
\usepackage{subcaption}
\hypersetup{hidelinks=true}
\usepackage{textcomp}
\usepackage{xcolor}
\usepackage[switch]{lineno}

\def\BibTeX{{\rm B\kern-.05em{\sc i\kern-.025em b}\kern-.08em
    T\kern-.1667em\lower.7ex\hbox{E}\kern-.125emX}}
\begin{document}
% \linenumbers
\title{AI-based prediction of worsening heart failure from low-resolution telemonitoring data}
\author{Erik~Aerts,~\IEEEmembership{}
        Yinan~Yu,~\IEEEmembership{}
        Annika~Rosengren,~\IEEEmembership{}
        Michael~Fu,~\IEEEmembership{}
        Martin~Lindgren,~\IEEEmembership{}
        Falk~Dippel, \\~\IEEEmembership{}
        Martin~Adiels,~\IEEEmembership{}
        and~Helen~Sjöland
\thanks{Manuscript submitted 18-12-2025.  This work was funded by the Wallenberg AI, Autonomous Systems and Software Program (WASP) funded by the Knut and Alice Wallenberg Foundation to YY, VINNOVA (Dnr. 2024-01446) to HS and YY, the Swedish state under the agreement between the Swedish government and the county councils regarding the education of physicians and research, the ALF agreement from the Healthcare Board, Region Västra Götaland (grants ALFGBG-991470, VGFOUREG-993916) to HS, and Forte, Sweden (grant 2024-00277) to HS.}
\thanks{E. Aerts and Y. Yu are with the Department
of Computer Science and Engineering, Chalmers Universiy of Technology, Gothenburg,
Sweden. Contact e-mail: aeerik@chalmers.se.}
\thanks{E. Aerts and Y. Yu are with the Department of Computer Science and Engineering, University of Gothenburg, Gothenburg, Sweden.}
\thanks{A. Rosengren, M. Fu, M. Lindgren, H. Sjöland are with the Departement of Molecular and Clinical Medicine, Sahlgrenska Academy, University of Gothenburg, Gothenburg, Sweden.}
\thanks{A. Rosengren, M. Fu, M. Lindgren, F. Dippel, H. Sjöland are with the Department of Medicine, Geriatrics and Emergency Medicine, Sahlgrenska University Hospital, Gothenburg, Sweden.}% <-this % stops a space
\thanks{H. Sjöland and Y. Yu is with the Center for Digital Health, Sahlgrenska University Hospital, Gothenburg, Sweden.}
%Center for Digital Health, vilken ligger under Sahlgrenska University Hospital.
\thanks{M. Adiels is with the Institute of Medicine, Sahlgrenska Academy, University of Gothenburg, School of Public Health and Community Medicine, Gothenburg, Sweden.}
\thanks{The code is available at: https://github.com/aeerik/TRACER.git.}}

\maketitle

\begin{abstract}
\textit{Objective}: Heart failure (HF) presents a healthcare challenge due to its high comorbidity burden, aging patient population and frequent hospitalizations. Remote monitoring offers a promising approach to managing HF patients by early detection of health deterioration. Developing autonomous systems to detect signs of worsening in telemonitoring data is of interest to reduce the workload of healthcare personnel. \textit{Methods}: We propose the TRACER model, a Transformer with Contrastive Event Representation, designed to predict timelines leading to rare hospitalization events in low-resolution and irregularly sampled telemonitoring data. TRACER incorporates time-aware embeddings for each biomarker, contrastive pre-training to enhance anomaly detection via representation learning, and independent binary classifiers for detection. We used measurement data containing remotely recorded biomarker sequences from 276 HF patients segmented into overlapping windows based on temporal rules, and labeled the windows based on the occurrence of HF relevant hospitalizations at the latter edge of the window. \textit{Results}: TRACER was able to correctly predict 66.7\% timelines leading up to HF hospitalizations in the highly imbalanced real-world dataset with an overestimation of 7.9 \%. Reformulating the training of TRACER as an event detection problem improved the predictive performance compared with training directly on forecasting windows, enabling more effective use of the limited hospitalization events. \textit{Conclusion}: TRACER demonstrated superior performance in detecting signs of worsening status in real-world telemonitoring data compared to the other tested models.  \textit{Significance}: TRACER shows promise in identifying signs of clinical deterioration that allow for alerts to be generated to provide counteractive treatment in patients with HF.
\end{abstract}


\section{Introduction}
\IEEEPARstart{H}{eart} failure (HF) is a complex syndrome caused by structural and/or functional abnormalities of the heart impairing its ability to pump blood effectively or to fill properly. This leads to inadequate circulation to meet the body's metabolic demands and triggers compensatory mechanisms such as hormonal and autonomic nervous system responses. HF is clinically defined by characteristic symptoms (e.g., breathlessness, fatigue and fluid buildup) and objective signs of cardiac dysfunction \cite{ESC2021}. The condition poses a significant burden on healthcare due to frequent hospitalizations, along with multiple comorbidities and old age \cite{Shahim2023Global, jones-2017-prognosis-heart-failure}. With increasing life expectancy in many countries, an ageing population, and the accumulation of multiple long-term conditions over the life course, these challenges associated with HF are expected to increase  \cite{Vigen2012HeartFailureAging}. \\
\indent To address this, efforts have been made to develop reliable systems for remote patient management \cite{Scholte2023TelemonitoringMeta}. Telemonitoring is one such solution, where continuous logging of physiological data, symptom scores and chat messages can assist remote health care personnel in detecting worsening health status. Remotely adjusting medications or performing proactive out-patient consultations before clinical worsening can prevent long and costly hospitalizations \cite{TeleMedMedicine}. The devices used for measuring information are often non-invasive, and the sampling of measurements can range in resolution from multiple measurements per second to single measurements per day \cite{Brahmbhatt2019RemoteHF, smartwatch}. The development of systems to detect clinical worsening and alert clinicians in telemonitoring data is of great interest, as such systems can help reduce the workload of healthcare personnel. Many existing telemonitoring systems tend to base clinical decisions primarily on the most recent set of recorded digital biomarkers, often without incorporating temporal context or trends. Another approach is threshold-based alerting, where notifications are triggered when specific values exceed predefined limits \cite{10783504, ANDRES201587}. Although this makes for easier detection of worsening in HF, sequential- and temporal dependencies present in the data will largely be ignored.  \\
\indent The integration of machine learning (ML) models into HF prediction tasks has accelerated over the past decade \cite{Habehh2021MLHealthcare}. However, shallow ML models often reduce the input space through feature selection, sacrificing potentially valuable information for simplicity and usability \cite{Dhal2022FeatureSelectionSurvey}. Deep learning (DL), a subfield of ML, is characterized by larger and more sophisticated neural architectures. DL models can achieve high performance by training on large datasets and have the capacity to incorporate multiple data modalities into the decision-making process \cite{LeCun2015DeepLearning}. Transformers have recently emerged as powerful architectures in DL \cite{Vaswani2017Attention}. Their self-attention mechanisms capture complex contextual relationships in sequential data, outperforming earlier models such as recurrent neural networks (RNNs) and Long Short-Term Memory (LSTM) \cite{RNN, Staudemeyer2019LSTM}.\\
\indent In this paper we present TRACER, a transformer-based DL model for predicting timelines leading to hospitalization, as a rare event representing worsening health in HF from low-resolution irregularly sampled telemonitoring data. The model is set up with transformer-based encoders using time-aware embeddings for each measured variable in the telemonitoring data, contrastive pre-training to enhance the ability to distinguish between normal- and anomaly patterns through representation learning and binary classification networks for each biomarker. We conduct several experiments to compare the viability of TRACER in various settings and the performance against other modeling options. Our contributions can be summarized as follows:

\begin{itemize}
    \item We show that using TRACER to forecast HF-related hospitalization, the model can benefit from training on a reformulated event detection task, enabling more effective use of the limited hospitalization events available in an imbalanced real-world telemonitoring dataset.
    \item TRACER successfully predicted 66.7\% of timelines leading to hospitalization events with an overestimation rate of  7.9\%.
    \item Evaluating TRACER shows the potential of using DL models to detect signs of clinical deterioration in order to generate alerts to clinicians in real-world telemonitoring data.
\end{itemize}

\section{Related work}
\subsection{Telemonitoring}
\noindent Telemonitoring is increasingly used for remote management of HF aiming to reduce frequent outpatient visits and hospitalizations by tracking patient-based health status in the hope of early detection of clinical worsening through home-based collection of health data \cite{Scholte2023TelemonitoringMeta}. Considerable efforts have been made to identify clinically feasible solutions, ranging between structured telephone support, non-invasive self-measurements, wearable devices and implantable sensors. Implantable sensor–driven care pathways represent an important and mature approach to remote HF management \cite{Helms2025}. For example, HeartLogic integrates data from heart sounds, thoracic impedance, respiratory patterns, night heart rate and physical activity into a composite index for early detection of worsening HF. These systems provide richer physiological information than vital-sign–based telemonitoring and may therefore offer higher specificity and clinical actionability in selected patients. However, their applicability is inherently limited to patients with appropriate indications for such implantable devices, typically for defibrillation/conversion or biventricular pacing. This represents a selected subset of the HF population, as device indications are mainly restricted to patients with persistently reduced left ventricular ejection fraction (LVEF) despite optimized therapy, sufficient life expectancy, and electrical dyssynchrony in the case of biventricular pacing \cite{ESC2021, McAlister2006CRTEligibility}. Garcia et al. conducted a multicenter prospective study in 310 HF patients with implantable cardioverter defibrillators, using weekly measurements of the HeartLogic index compared to a static threshold to trigger clinician assessment over a 12 month follow-up period \cite{Garcia2025}. Boehmer et al. conducted the PREEMPT-HF study, a multicenter prospective observational trial including 2183 HF patients with implantable cardioverter defibrillators or cardiac resynchronization therapy devices. The study showed that implantable sensor trends, including the HeartLogic Index, exhibit systematic changes in the period surrounding clinical events related to HF \cite{boehmer2023}. \\
\indent For broader application of monitoring in HF patients without the invasive procedures, non-invasive and user-friendly approaches are required  \cite{tersalvi2021}. A large meta-analysis of mortality and rehospitalization outcomes also reported more favorable results for non-invasive than invasive monitoring, firmly supporting a role for non-invasive telemonitoring in remote HF management \cite{Scholte2023}.  Stehlik et al. evaluated a multisensor wearable patch for up to three months in patients recently hospitalized with HF. Using machine learning, the system detected impending rehospitalization a median of 6.5 days before the event \cite{stehlik2020}. Llussa et al. applied an ML model to a telemonitoring platform to reduce the number of false positive alerts generated by a rule-based threshold system using measurements from wearable devices \cite{10783504}. \\
\indent While sensor-based and wearable telemonitoring approaches enable detailed physiological monitoring, many systems require dedicated devices or infrastructure which may limit their scalability and widespread deployment in routine HF care \cite{scholte2024}. Patient-related barriers include limited battery life, discomfort, restrictions on showering or movement, device bulk, and skin irritation during prolonged use. Staff-related barriers include poor workflow integration and the need to review and interpret large volumes of incoming data \cite{CAJITA2026619}. Fewer studies have investigated predictive models for telemonitoring systems based on routinely collected physiological measurements. An early example by Koulaouzidis et al. modeled daily measurements of heart rate, blood pressure, and weight from 308 HF patients as discrete-time signals and used multiresolution analysis to extract predictive patterns across different temporal scales. Statistical features derived after high-pass and low-pass filtering were subsequently used with a Naïve Bayes classifier to predict imminent HF hospitalizations \cite{KOULAOUZIDIS201678}.

\subsection{Medical time series}
\noindent Medical time series is a subdomain of medical data where temporal spacings between recordings are used as a source of information \cite{Glass1993}. Similar to time series data in a general domain, medical time series contain two data levels as a baseline: sample and observation levels. However, medical time series often carry additional levels of information, such as which patient or trial the medical time series originates from \cite{wang2023contrasteverythinghierarchicalcontrastive}. The frequency of sampling in medical time series varies depending on the source and task: from multiple recordings per second to spaced measurements taken with daily- or weekly intervals. Resulting datasets are therefore characterized by either high- or low-resolution medical time series \cite{singh2019multiresolutionnetworksflexibleirregular, Brahmbhatt2019RemoteHF, smartwatch}. Examples of high-resolution medical time series include continuously recorded electrocardiograms (ECG) and electroencephalograms (EEG), which capture detailed cardiac and neurological activity at fine temporal scales. In contrast, low-resolution medical time series include checkup tests of biochemical markers, such as thyroid and liver function tests conducted every six months to monitor systemic effects during amiodarone treatment \cite{amidarone}. \\
\indent The sampling frequency of measurements in medical time series is often not uniform. The resulting characteristics of the irregularly sampled medical time series require additional considerations when selecting suitable methods of modeling. Models for irregularly sampled medical time series can be divided into two categories depending on how the data is handled: (1) treating sampling irregularity as missing data which requires sophisticated imputation techniques; (2) modeling the irregular time spacings as a source of information. Previous work has utilized both approaches, and for the case of modeling the irregular time spacings, adapted RNNs have been employed \cite{singh2019multiresolutionnetworksflexibleirregular, baytas2017patient}. Baytas et al. introduced a Time-Aware LSTM (T-LSTM) which modifies the standard LSTM cell to incorporate elapsed time between observations \cite{baytas2017patient}. The model applies a decay mechanism to the memory cell which reduces the influence of past information as time gaps increase. This has been applied to tasks such as patient subtyping in Parkinson’s Progression Markers Initiative dataset \cite{baytas2017patient}. More recently, Liu et al. proposed a framework that combines a Transformable Time-aware Convolution Network with an Irregular Fourier Analysis Network to encode temporal and spectral representations from irregularly sampled physiological data for time series forecasting. While such methods have demonstrated promising performance on retrospective forecasting tasks using critical care datasets, their applicability to low-resolution telemonitoring for rare event prediction remains less explored \cite{liu2026}.

\subsection{Contrastive learning}
\noindent Contrastive learning trains effective representations of the data by arranging similar data pairs close together whilst simultaneously pushing dissimilar data pairs further apart in feature space \cite{tian2020makesgoodviewscontrastive}. The interpretation of what should be considered as data pairings can be designed based on the task at hand \cite{wang2023contrasteverythinghierarchicalcontrastive}. Moreover, the pairs can be formed from multiple data modalities making the approach particularly powerful for cross-modal understanding and retrieval tasks \cite{radford2021learningtransferablevisualmodels}. \\
\indent Previous work based on contrastive learning has utilized many different data sources and data modalities. Chen et al. published the framework SimCLR for contrastive visual representation learning \cite{chen2020simpleframeworkcontrastivelearning}. The framework quickly became a popular choice for contrastive learning methods for vision tasks due to high recorded performance and simplicity \cite{zhang2024understandingbenefitssimclrpretraining}. SimCLR uses augmented views of the same image to create data pairings to form the self-supervised learning framework \cite{chen2020simpleframeworkcontrastivelearning}. SimCLR focuses on learning visual representations through augmentations of the same image, whereas CLIP introduced by Radford et al. extends contrastive learning to a multi-modal setting. CLIP jointly trains an image- and text encoder to project images and their corresponding textual descriptions into a shared embedding space. The model treats related images and texts as data pairings, learning semantic representations between the modalities \cite{radford2021learningtransferablevisualmodels}. The COMET framework by Wang et al. is a contrastive framework built for medical time series. COMET uses four different contrastive blocks for the observation, sample, trial and patient data levels in medical time series. The contrastive blocks form different data pairings depending on the respective data levels \cite{wang2023contrasteverythinghierarchicalcontrastive}. Another domain-specific adaptation of contrastive learning is the CARLA framework by Darban et al. for anomaly detection in time series data.  The CARLA framework augments time series data with injected anomalies and forms data pairings between time windows either both- or neither carrying an injected anomaly. With this, CARLA encourages the learning of discriminative representations that effectively separate normal and anomalous patterns in time series data \cite{Darban_2025, ECG, EEG}. \\

\noindent The present approach is designed for non-implanted patients and relies on lower-cost, broadly deployable monitoring that can be implemented at scale, including in recently diagnosed patients, older multimorbid patients, and healthcare settings where implantable diagnostics are unavailable or not indicated. Certain previous work on autonomous systems for detecting adverse events in HF telemonitoring data has relied on sensor-based data or the most recent recording for assessment \cite{10783504, stehlik2020}. However, accessible measurement-based telemonitoring data contain temporal and hierarchical information that can be utilized, including irregular sampling patterns, variable measurement resolutions and patient profiles.  Recent approaches have investigated methods for modeling irregularly sampled medical time series through temporal and spectral representations \cite{baytas2017patient, liu2026}. However, the application of such approaches to low-resolution telemonitoring data for rare event prediction remains less explored. Our proposed model leverages the temporal structure and patient-level information to learn patterns in telemonitoring data and employs contrastive learning objectives to enhance representation learning for predicting timelines leading to adverse events. In this work, we aim to showcase the potential of the model to utilize low-resolution medical time series with irregular sampling to predict timelines for adverse events, even under the high class imbalance characteristic of real-world clinical data.

 \section{Telemonitoring data}
\begin{figure*}[t]
    \centering
    \includegraphics[width=0.8\linewidth]{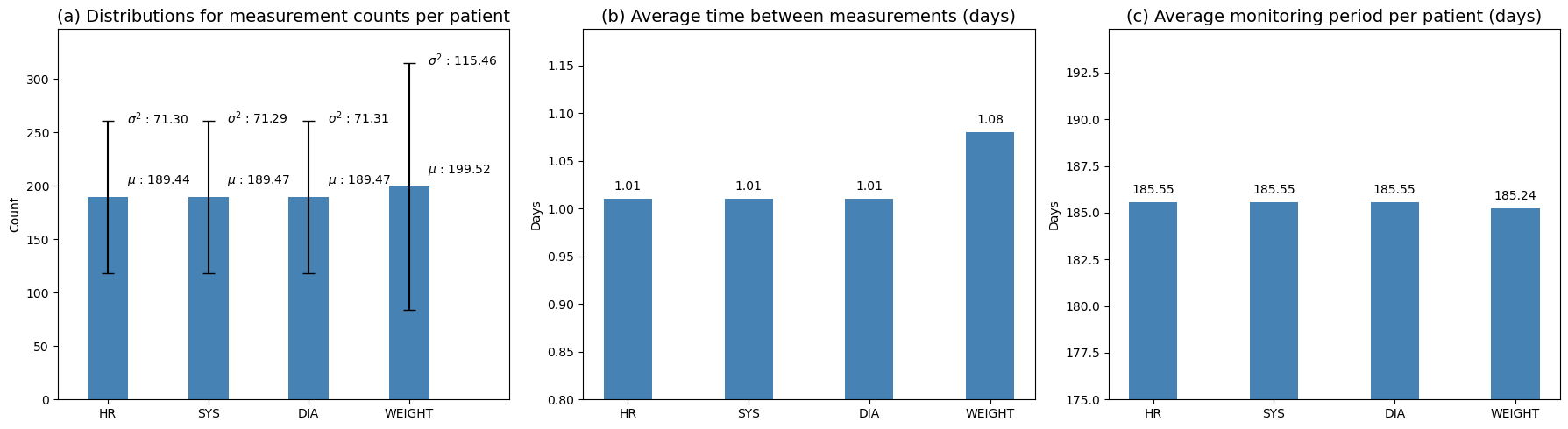}
    \caption{Data overview for monitored biomarkers in the telemonitoring. From the left: (a) mean ($\mu$) and standard deviation ($\sigma^2$) of measurement counts per patient; (b) the average number of days between measurements; (c) the average number of days in the monitoring period per patient  }
    \label{fig:physiometric}
\end{figure*}

\begin{figure}
    \centering
    \includegraphics[width=0.9\linewidth]{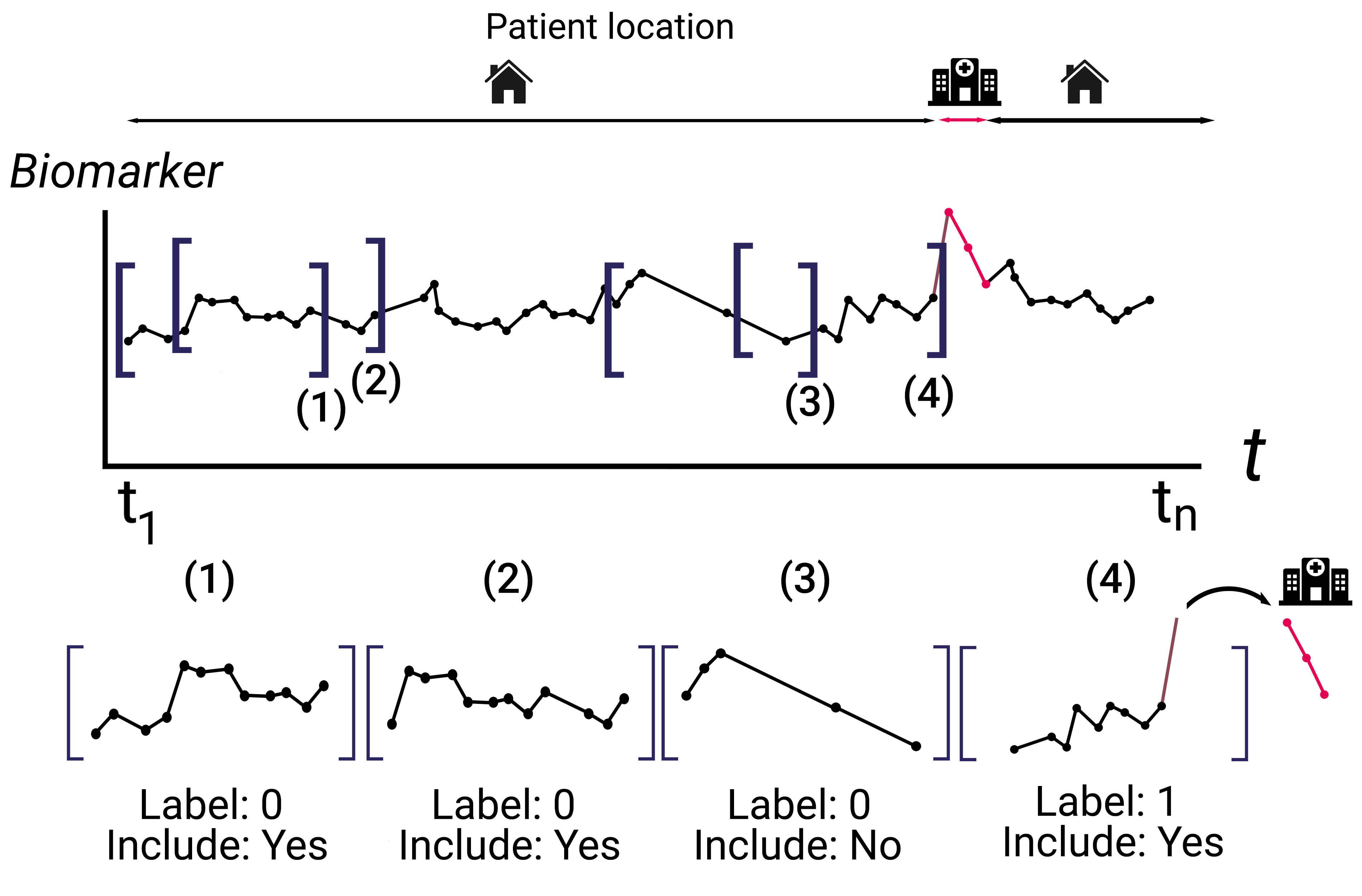}
    \caption{ Illustration of the window process. Four windows from the recorded variables measurement signal for a patient (top) are highlighted with bars, and the location of the patient during the period is showcased with black (home) and red (hospital). Windows 1  and 2 show windows created from an overlapping region of measurements. One out of four windows does not contain sufficient data points and is therefore discarded (window 3). One out of four windows contains a hospital event (red) at the latter edge and is therefore labeled with label 1 accordingly (window 4).}
    \label{fig:windowprocess}
\end{figure}

\begin{figure}
    \centering
    \includegraphics[width=0.9\linewidth]{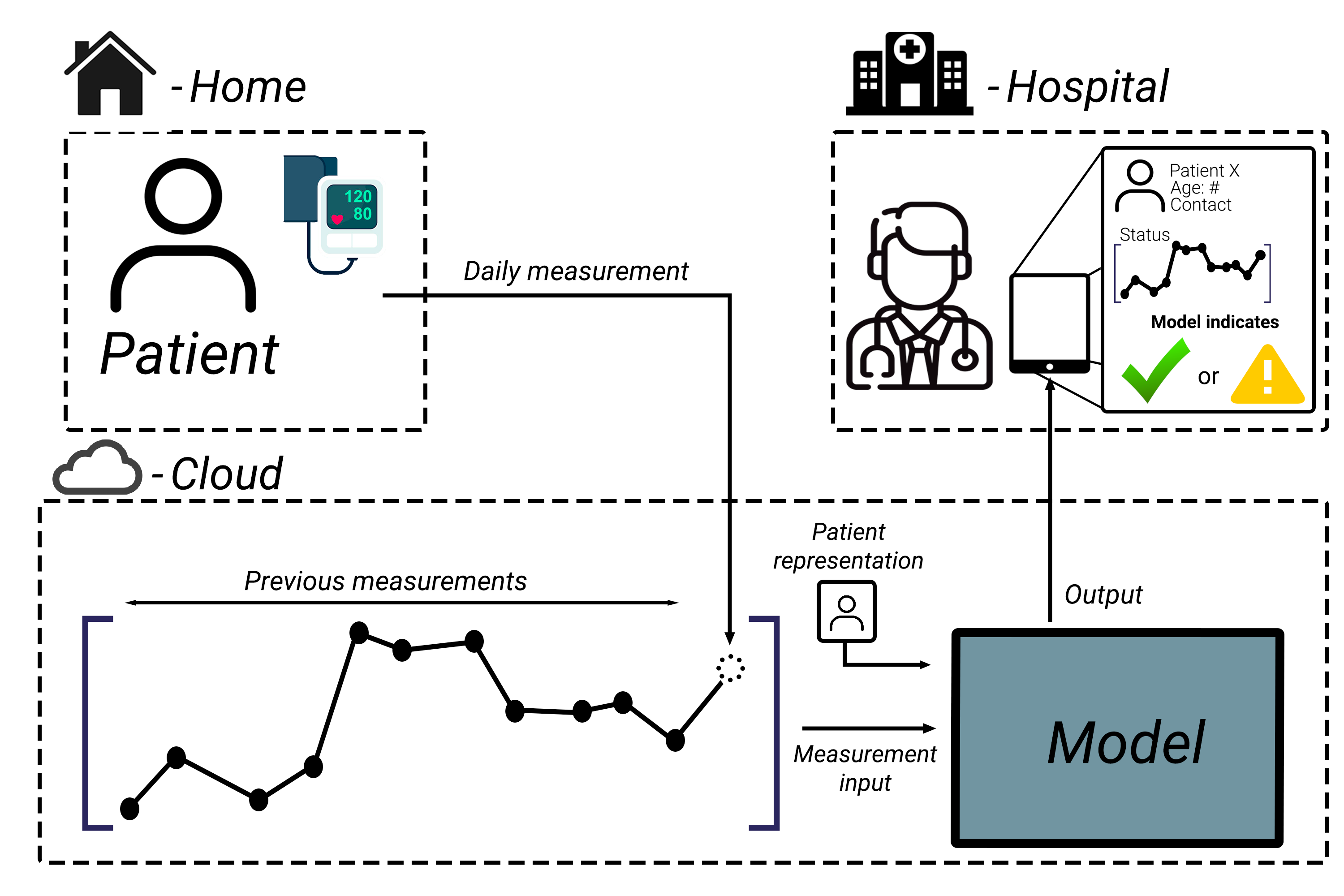}
    \caption{ Illustration of telemonitoring pipeline. The patient at home measures daily recordings of their biomarkers with the provided devices (top left). The data is joined with previous recordings and passed, alongside a representation of the patient profile, to the DL model hosted on a cloud-based service (bottom). The output of the model is sent to the hospital (top right) to provide an indication of the health status of the patient to clinicians. }
    \label{fig:pipeline}
\end{figure}

\noindent Patients with new or worsening HF judged to be in clinical need of specialist cardiology follow-up  were included for participation in a multicohort telemonitoring study conducted across three hospitals within Region Västra Götaland, Sweden. The study was approved by the Swedish Ethical Review Authority, Stockholm Department 4 (Etikprövningsmyndigheten, Sverige; Dnr 2020-02632). Each patient was monitored for 6 months after receiving the equipment for measurements during the study period: 2020-06-01 to 2023-05-31. The participants were tasked to perform daily measurements of biomarkers, which were digitally recorded and delivered by wireless transmission to the care provider for evaluation by a nurse. The recorded biomarkers were heart rate (HR), systolic- and diastolic blood pressure (SYS, and DIA respectively) and body weight. The patients were included in the study if (1) one of the following applies: newly diagnosed HF requiring up-titration of medication; HF with instability with at least one hospital admission due to HF in the past 12 months; HF with instability;  HF requiring up-titration/optimization of medication in preparation for a potential intervention. (2) all of the following apply: understands spoken and written Swedish (with assistance if needed); cognitive and physical ability to manage remote monitoring (with assistance if needed); has online identification; owns a smartphone or tablet; has internet access; comorbidity risk assessment completed with no comorbidity expected to negatively affect monitoring, consented to be a part of the study. Early withdrawal from the study or longer absence from monitoring due to critical worsening of health status or other personal reasons were recorded. Data were collected and separated into three categories: (1) baseline patient data at inclusion. Baseline data contained demographic features, physiological information, clinical history and current medications with respective dosages at the time of inclusion; (2) recorded measurements for the HR, SYS and DIA variables during the telemonitoring period with measurement value and a timestamp. However, patient body weights were inconsistently recorded compared to the other biomarkers. The measurement counts per patient for weight had a high variance (Fig. \ref{fig:physiometric}a), indicating a patient bias in the data resulting in a higher average time between measurements (Fig. \ref{fig:physiometric}b) and lower average monitoring period (Fig. \ref{fig:physiometric}c). Therefore, weights were excluded from the model development to maintain data consistency; (3) A complete list of all hospitalizations between the inclusion date of the first patient and the completion date for the last patient. One patient died during the study period due to causes unrelated to the scope and objectives of the study. The measurement data of the patient were included in the study. All model development and evaluation were performed retrospectively using previously collected telemonitoring and hospitalization data. Model outputs were not available to the clinical team during patient monitoring and therefore did not generate alerts or trigger any clinical assessments, treatment changes, or other actions. \\
\indent Initial preprocessing of the measured data was based on physiological plausibility. Outliers from human or system errors in the measurement data were excluded based on clinically validated ranges for each biomarker. Patient records were examined by a clinician for classification of each hospitalisation and its relevance for HF, labeling each event as either clinically relevant or non-relevant event. The sequential measurement data were divided into shorter data intervals, denoted as windows. The windows were created using a set of spatial and temporal rules (Fig. \ref{fig:windowprocess}). Given a starting measurement point $m_{i}$, a set of $s$ measurement points ranging from $m_{i}$ to the acceptable time range of 21 days in advance was created  $s \in (m_{i},m_{i+n})$. Measurement points were then added to the window in sequential order from the available set $s$ until the window contained 14 measurement points. If enough measurement points were not found in $s$ to create a window of at least a length of 10, the window was discarded. The initial starting point $m_{i}$ was then shifted a percentage of the window length forward, creating a window overlap, and the process was repeated. The window size was selected based on the telemonitoring protocol. A maximum window length of 14 measurements was chosen to represent approximately two weeks of monitoring, while allowing flexibility to account for irregular sampling in the collected data. If the end of the dataset was reached and an insufficient number of measurement points were left in $s$, the start of $s$ was expanded with earlier measurement points to reach the minimum threshold. Each created window was assigned a binary label of 1 if a hospitalization relevant for HF was present for the respective patient directly after the latter edge of the window, and 0 otherwise. "The latter edge" refers to the timestamp of the last measurement within the window, with a hospitalization occurring within 24 hours after this last recorded measurement defining a positive label. This formulation was designed to evaluate whether measurements preceding a hospitalization event contained temporal patterns associated with increased risk of deterioration. Thus, the model was trained to identify timelines leading to observed HF-related hospitalizations rather than concurrent hospitalization states. The hospitalization labels were derived from clinical records. Lastly, both the measurement values and the timestamps in each window were normalized. To maintain sequence consistency, all windows were padded to reach an input length of 14. The labeled windows were used as input to the DL models. Given the short window of the telemonitoring time series, the models were tasked with predicting a binary label for each window, indicating the patients clinical status. \\
\indent The envisioned deployment scenario of the DL model would operate as a support tool within the existing clinician guided HF telemonitoring pipeline. Model outputs would be reviewed by a clinician or nurse to alert to possible deterioration to motivate intensified patient contact, additional assessment or adjustment of follow-up checkups. An overview of the envisioned telemonitoring pipeline with an integrated DL model can be seen in Figure \ref{fig:pipeline}.


\section{Methods}
\subsection{Deep learning model}
\begin{figure*}[t]
    \centering
    \includegraphics[width=0.8\linewidth]{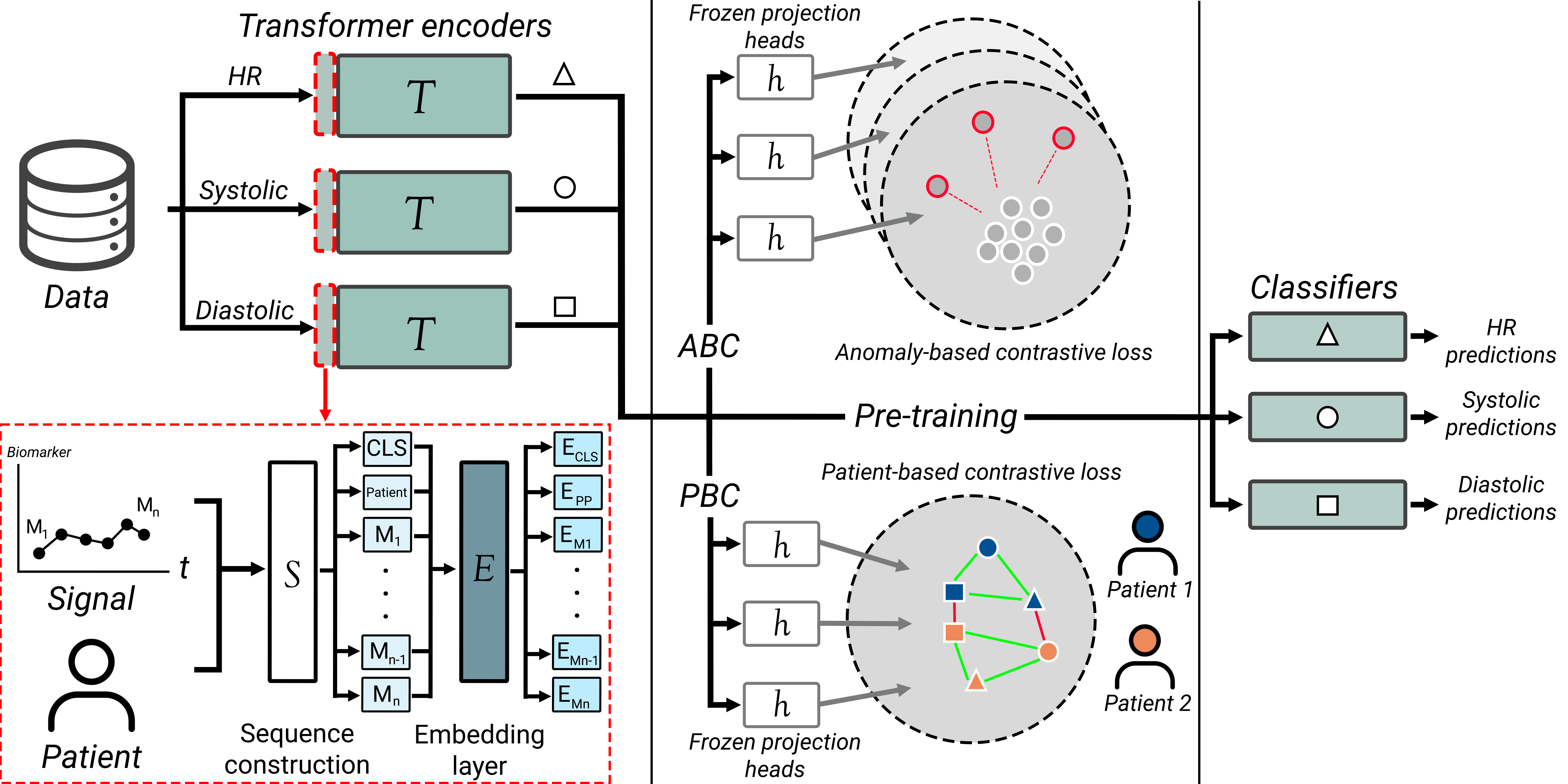}
    \caption{Illustration of the model architecture. The model is divided into three parts: transformer encoders (left), pre-training (middle) and classification networks (right). The embedding process is shown in the red-striped box. The tokenized measurement sequence is shown as boxes labeled $M_{i}$ and embeddings are shown as boxes labeled $E_{i}$. Each biomarker is illustrated as a different geometrical shape as the output from the encoders. For ABC pre-training (top), each representation is projected into a specific feature domain where the hospitalized windows (red) are separated from the non-hospitalized windows (white). For PBC pre-training (bottom), all representations are projected into the same domain where each window from the same patient is regarded as a positive pair (green link) and each window from different patients is seen as a negative pair (red link). \\ 
    \textit{Abbreviations: t = Time, HR = Heart Rate, SYS = Systolic Blood Pressure, DIA = Diastolic Blood Pressure, M = Measurement} }
    \label{fig:model}
\end{figure*}

 \noindent We built a model architecture that consisted of three parts: (1) parallel biomarker-specific transformer encoders; (2) patient-based contrastive (PBC) pre-training and anomaly-based contrastive (ABC) pre-training; (3) supervised binary classification networks (Fig. \ref{fig:model}). The model used three encoders, one for each biomarker in the study, which used self-attention to understand the contextual relationships between the different embeddings in a given sequence. Each encoder was built with a set of attention blocks. The attention blocks were constructed with four parallel attention heads, a technique in which the input is projected into multiple subspaces simultaneously to compute parallel attention functions independently. Following this, an add\&normalize layer (A\&N) summed the input- and output embeddings from the previous layer and normalized the sum. Continuing, a position-wise feed-forward layer (PW-FFN) was applied to each position in the input sequence. Lastly, an additional A\&N for the PW-FFN layer concluded the attention block. In total, each encoder consisted of three attention blocks. The output embeddings from the encoders were passed to a secondary set of networks depending on the training schedule. Two types of pre-training were available before classification of the embeddings. During classification, the enriched embeddings were passed to a set of classifying networks. These classifying networks were built as a two-layered fully connected network resulting in a binary output determining if the model predicted hospitalization or not. 

 \subsection{Contrastive pre-training}
\subsubsection{Patient-based contrastive}
During PBC pre-training, the encoders passed on the enriched embeddings to a set of projection heads $H$. These projection heads were two-layered neural networks that transformed the embeddings $x_i$ to a lower-dimensional representation $z_i = H(x_i) $. The projection heads were frozen to ensure that the encoder adapted its representations to the task constraints. The transformed embeddings were then projected into a structured space, where similarity between transformed embeddings was evaluated. To accommodate varying batch sizes between biomarkers, the similarity matrices were padded and a sharpened sigmoid function was then used to modulate these similarity scores by emphasizing the distinction between highly similar and dissimilar pairs. The pairing was determined by calculating the cosine similarity between embedded versions of the patient profiles $\mathbf{p}$. If the patient profile of two windows, either from the same or two different biomarker, exceeded a similarity threshold $\theta$ greater than- or equal to 99\%, the windows were determined to be a positive pair. The PBC loss $\mathcal{L}_{\text{PBC}}$ for the window representations $z$ was defined as:

% Now start a new group and environment on the next line
{\small
\begin{align}
\mathcal{L}_{\text{PBC}} &= \frac{1}{N(N-1)} \sum_{\substack{i=1}}^{N} \sum_{\substack{j=1 \\ j \neq i}}^{N} \Bigg[
y_{ij} \cdot \log \left( \sigma \left( \tau_p \cdot \frac{\mathbf{z}_i^\top \mathbf{z}_j}
{\|\mathbf{z}_i\| \cdot \|\mathbf{z}_j\|} \right) \right) \notag \\
&\quad + (1 - y_{ij}) \cdot \log \left( 1 - \sigma \left( \tau_p \cdot \frac{\mathbf{z}_i^\top \mathbf{z}_j}
{\|\mathbf{z}_i\| \cdot \|\mathbf{z}_j\|} \right) \right)
\Bigg] \\
y_{ij} &= \mathds{1} \left\{ \frac{\mathbf{p}_i^\top \mathbf{p}_j}{\|\mathbf{p}_i\| \cdot \|\mathbf{p}_j\|} \geq \theta \right\} \notag
\end{align}
}

\noindent where $N$ denoted all available lower dimensional window representation $z$. Moreover, $\tau _p$ was a trainable temperature parameter to sharpen the distance calculation for the sigmoid function $\sigma$. The $ \mathds{1}$ represented the indicator function setting $ y_{ij} = 1$ if the cosine similarity between profile vectors $\mathbf{p}_i $ and $\mathbf{p}_j$ was greater than or equal to the threshold $\theta$, otherwise 0.
\subsubsection{Anomaly-based contrastive}
During ABC pre-training, we used a similar setup of frozen, two-layered projection heads $H$  to transform the output embeddings  from the encoder $x_i$ to a lower dimensional representation $z_i = H(x_i) $. The embedding representations were projected into structured spaces to compute similarity relationships. The positive- and negative pairings were formed based on the window labels $\ell$. The model computed the cosine similarity of two representations to assess their similarity, which aimed to push positive pairs together and negative pairs further apart in the space through a sharpened sigmoid function. The ABC loss $\mathcal{L}_{\text{ABC}}$ for the window representations $z$ was defined as:

{\small
\begin{align}
\mathcal{L}_{\text{ABC}} &= \frac{1}{N(N-1)} \sum_{\substack{i=1}}^{N} \sum_{\substack{j=1 \\ j \neq i}}^{N} 
\Bigg[
y_{ij} \cdot \log \left( \sigma \left( \tau_a \cdot \frac{\mathbf{z}_i^\top \mathbf{z}_j}
{\|\mathbf{z}_i\| \cdot \|\mathbf{z}_j\| \cdot \gamma} \right) \right) \notag \\
&\quad + (1 - y_{ij}) \cdot \log \left( 1 - \sigma \left( \tau_a \cdot \frac{\mathbf{z}_i^\top \mathbf{z}_j}
{\|\mathbf{z}_i\| \cdot \|\mathbf{z}_j\| \cdot \gamma} \right) \right)
\Bigg] \\
y_{ij} &= \mathds{1}\left\{ \ell_i = \ell_j \right\} \notag
\end{align}
}
\noindent where $N$ denoted all available lower dimensional representation $ z$. The $\gamma$ parameter represented a set scaling parameter for the cosine similarity and $\tau _a$ was a trainable temperature parameter to sharpen the distance calculation for the sigmoid function $\sigma$. The $ \mathds{1}$ represented the indicator function setting $ y_{ij} = 1$ if the labels $\ell_i$ for lower-dimensional window representation $z_i$ and $\ell_j$ for lower-dimensional window representation $z_j$ were equal, and 0 otherwise.

 \subsection{Embedding}
\begin{figure}[H]
    \centering
    \includegraphics[width=1\linewidth]{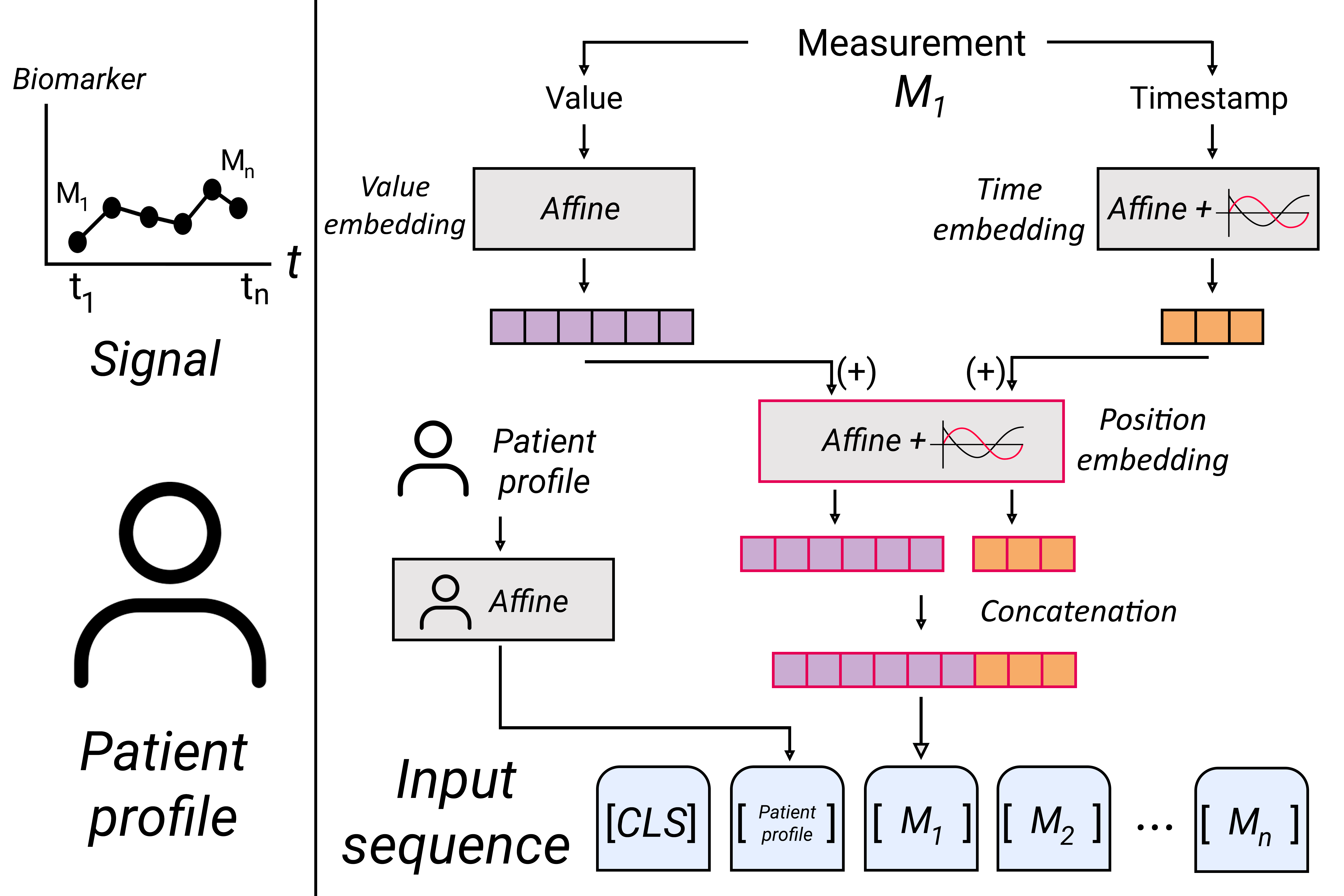}
    \caption{Illustration of the embedding process. The sources of information are shown to the left. Measurement $M_1$ in the figure is decomposed into a value and a timestamp. The value undergoes affine transformation to produce a value embedding, whereas the timestamp is converted into a time embedding with sinusoidal encoding. These are then further transformed to form position-aware embeddings followed by concatenation. The patient profile is independently embedded through an affine layer. All components: a \textit{CLS}-token, the patient profile, and the measurement embeddings are concatenated to form the input sequence. \\
    \textit{Abbreviations: t = time, M = measurement }}
    \label{fig:embedding}
\end{figure}

 \noindent The embedding system consisted of three modules: (1) value embedding; (2) time embedding; (3) positional embedding. The measurement values were embedded with a linear layer projecting the value to a higher-dimensional space of size $d_e$. Timestamps were embedded using sinusoidal encoding where each relative UNIX time was mapped to a higher-dimensional space of size $0.5 d_e$ using an affine transformation followed by a periodic function. To maintain the sequential order information in the input, absolute positional embeddings based on sinusoidal functions were added separately to the value and time embeddings. Although the applied encoding is absolute, this design enables the attention mechanism to learn relative relationships between measurements. The value- and time embeddings were then concatenated to create a joint embedding representation for each measurement. A patient profile token was added to the measurement sequence by using a linear layer to project the patient profile information to a higher-dimensional space of size $1.5d_e$ to maintain dimensional consistency with the joint embeddings. Finally, a \textit{CLS}-token with random initialization 
was added at the start of each sequence (Fig. \ref{fig:embedding}).

\subsection{Comparison models}
\noindent A gradient-boosting machine learning model was implemented as a comparison to the proposed transformer-based model architecture in the project. XGBoost, a parallel tree-boosting model, used a maximum number of trees set to 200 and a maximum depth of 7. To maintain temporal elements in the input, parameters from a window-specific autoregressive integrated moving average (ARIMA) model were added as features. The ARIMA models for all windows were fitted on the previous, current and next time windows for stability. In cases where the ARIMA model failed to converge or generate valid parameters, the corresponding window was excluded from the ARIMA/XGBoost evaluation. The \textit{p}, \textit{d} and \textit{q} parameters for each ARIMA model were added as features alongside measurement values for the windows. The XGBoost model was fitted with a learning rate of $10^{-4}$ with class weights depending on the imbalance ratio of the dataset similar to the transformer model. The same preprocessing steps of the windows were performed for the ARIMA/XGBoost model.  \\
\indent  Two LSTM recurrent neural networks were set up as secondary comparison models. The first LSTM model was implemented as a four-layer bidirectional network with a hidden state dimension of 32 per direction. Following the recurrent layers, mean pooling was performed across all time steps to aggregate information into a fixed-size representation. This representation was passed through a fully connected output layer to perform binary classification. Each biomarker was processed in parallel to one another from separate LSTM models and the predictions were concatenated before evaluation. The input window to the LSTM had two dimensions:  the measurement values and their respective normalized time spacing to the start of the window to maintain the temporal features in the data. Secondly, a T-LSTM model based on the approach proposed by Baytas et al. \cite{baytas2017patient} was implemented as an additional recurrent baseline. The model incorporated elapsed time between consecutive measurements through a decay mechanism applied to the short-term component of the cell state. The T-LSTM processed biomarker measurements together with their corresponding time gaps and used the final hidden state as a patient representation for binary classification through a fully connected prediction head. Both LSTM based models were trained for a maximum of 40 epochs with a learning rate of $10^{-4}$, with class weights depending on the imbalance ratio of the dataset like previous models. 

% An overview of the total parameter counts for the models used in the study is presented in Table \ref{tab:params}.

% \begin{table}[H]
% \centering
% \caption{Overview of the parameter counts for the models used in the study.}
% \renewcommand{\arraystretch}{1.2} % Optional: improves spacing between rows
% \begin{tabular}{p{3cm}p{3cm}}
% \hline\hline
% Model architecture & Parameter count  \\
% \hline

% \quad Transformer & 25837412 \\
% \quad LSTM & 253635 \\
% \quad TLSTM & 17461599 \\
% \quad XGBoost & Up to 153000 nodes \\

% % \quad \% windows with with $<$ 12 hour gap & 46.97  \\
% % \quad \% windows with with $>$ 36 hour gap & 34.41  \\

% \hline\hline
% \end{tabular}
% \label{tab:params}
% \end{table}

\section{Experiments}
\subsection{Patient overview}
\noindent A total of 276 patients with new- or worsening HF were monitored, including n=72 women (26.1\%) with a mean age of 67.6 $\pm$ 12.18 years. The largest proportion of patients were enrolled due to newly diagnosed HF requiring up-titration of medication (n=224, 81.1\%), followed by HF with instability (n=27, 9.8\%) and with at least one hospital admission due to HF in the past 12 months (n=25, 9.1\%). No patients were enrolled due to HF requiring up-titration/optimization of medication in preparation for a potential intervention.  The majority of patients had HF with reduced ejection fraction (HFrEF, n=225, 81.5\%), while fewer patients had HF with mildly reduced ejection fraction (HFmrEF, n=41, 14.9\%) or preserved ejection fraction (HFpEF, n=10, 3.6\%). During monitoring, 103 hospital admissions were recorded from 56 individuals, of which 20 admissions from 18 individuals were classified as relevant for the HF condition. This resulted in an imbalance ratio of 1244.63 between the two classes, which decreased to 502.51 through up- and downsampling in the training set. 

\subsection{Experimental setup}

\noindent An overview of the properties of the data windows used in the training is shown in Table \ref{tab:windowtable}. The dataset was split into training and validation sets with a 90/10 ratio on a patient-independent basis, ensuring no patient contributed data to both sets. To counteract class imbalance, upsampling and downsampling were applied to the training set, by inserting duplicate hospitalization windows randomly within the same patient’s sequence and discarding 20\% of non-hospitalization windows, respectively. No resampling was applied to the validation set; all reported metrics were computed on the original class distribution. To address the low number of events in the dataset, the split was reshuffled until at least four patients with hospitalization events appeared in the validation set. Windows were created with 75\% overlap, and an initial batch size of 32 was used. Due to varying measurement counts across biomarkers, batch sizes and counts were adjusted to ensure consistent batch numbers. The model used a joint embedding dimension of 384, and a projection dimension of 64 for similarity comparisons. A dropout rate of 10\% was applied throughout training. The learning rate was fixed at $10^{-6}$ for both pre-training (30 epochs) and fine-tuning (100 epochs), without early stopping. Binary cross-entropy loss with class weights reflecting class imbalance was used, optimized with Adam and weight decay. Initial sigmoid sharpness parameters, $\tau_p$ and $\tau_a $, were set as 5 and 3 respectively and the scaling parameter $\gamma$ was set to 0.9. \\
\indent We evaluated model performance using sensitivity and specificity for accurately predicting timelines leading to hospitalization or not. Moreover, we evaluated the expected clinical alert burden in case of model use by calculating false alerts per patient-month (FA-PPM) and the number needed to evaluate (NNE). We defined FA-PPM as the ratio between the total number of false positive alerts divided by the total monitoring time across the cohort, and the results are presented in all applicable tables:
\begin{equation}
    \text{FA-PPM} = \frac{\text{False positive alerts}}{\text{Number of patients} \cdot \text{Number of months} }
\end{equation}
Moreover, we defined NNE as the ratio between the total number of positive alerts generated and the number of correctly detected hospitalization events:
\begin{equation}
    \text{NNE} = \frac{\text{True positive alerts} + \text{False positive alerts}}{\text{True positive alerts} }
\end{equation}
\begin{table}[H]
\centering
\caption{Overview of the properties of the constructed time windows of measurements in both datasets. All values presented in the table are calculated without up- or downsampling.}
\renewcommand{\arraystretch}{1.2} % Optional: improves spacing between rows
\begin{tabular}{p{6cm}p{2cm}}
\hline\hline

\quad Number of windows & 51030 \\
\quad Number of non-hospital windows & 50989 \\
\quad Number of hospital windows & 41 \\
\quad Average window length (measurements) & 13.58 \\
\quad Average delay between measurements (days) & 0.957 \\
% \quad \% windows with with $<$ 12 hour gap & 46.97  \\
% \quad \% windows with with $>$ 36 hour gap & 34.41  \\

\hline\hline
\end{tabular}
\label{tab:windowtable}
\end{table}

\subsection{Main results}
\subsubsection{Event-based training for timeline prediction}
\noindent  We compared alternative training configurations and model architectures to investigate how well the models were able to predict timelines leading to hospitalization in the validation data. Moreover, we used various combinations of the pre-training techniques for the transformer model to gauge what impact the techniques had on the performance. Each model was trained on a reformulated event-based detection task in which hospitalization events were not fixed at the latter edge of the time windows, but could occur at any point within it. With window overlap, this formulation exposed the models to multiple different views of the few hospitalizations present in the training set which aimed to enrich the models understanding of the contextual relationships around the events (Table \ref{tab:DecFor}). The transformer model without pre-training scored higher on sensitivity compared to the LSTM- and ARIMA/XGBoost models, whilst having a similar rate of overestimation. All models generated a comparable FA-PPM score of approximately two to three false alerts per patient and month. However, the NNE differed between models, ranging from 15.9 for the T-LSTM model to 286.5 for the PBC pre-trained transformer. Because of the class imbalance and subsequent rarity of the hospitalization events, even modest differences in the number of correctly detected hospitalizations translated into substantial differences in the number of clinical evaluations required to identify one true hospitalization. \\ 
\indent Using certain pre-training increased the performance of the transformer model. By using only the ABC pre-training, the ability to accurately predict timelines leading to hospitalizations increased by 3.8 percentage units whilst decreasing the hospitalization overestimation with 1.1 percentage units compared to not pre-training the transformer. This model architecture was named the TRACER (Transformer with Contrastive Event Representation) model. Using the PBC pre-training resulted in a substantially decreased performance compared to not pre-training the model. When both pre-training techniques were applied in succession, the order in which they were used affected model performance. Applying PBC pre-training before ABC led to a better ability to detect timelines leading to hospitalization, compared to the alternative order which achieved a lower overestimation. Both combinatorial setups of pre-training techniques for the transformer achieved a lower sensitivity to TRACER with similar- or lower specificity, indicating that excluding PBC from the pre-training was beneficial for the task.
% \begin{table}[H]
% \centering
% \caption{Overview of the performance of predicting timelines leading to hospitalization for the tested architectural setups trained on event-based detection in the study.   }
% \begin{tabular}{lllll} \hline\hline
% Model type (Pre-training) & Sensitivity &Specificity   & FA-PPM & NNE  \\
% \hline
% Transformer (No pre-train) & 0.629 & 0.910 & 2.86 & 24.87 \\
% TRACER & 0.667 & 0.921 & 2.49 & 20.79\\
% Transformer (PBC) & 0.037 & 0.937 & 2.01 & 286.53 \\
% Transformer (PBC\&ABC) & 0.518 & 0.909 &  2.86 & 30.14 \\
% Transformer (ABC\&PBC) & 0.370 & 0.935 & 2.04 & 30.00 \\
% LSTM & 0.480 & 0.932 & 2.14 & 16.12  \\
% T-LSTM & 0.520 & 0.926 & 2.33 & 15.87 \\
% ARIMA/XGBoost & x & x & x \\
%  \hline\hline
% \end{tabular}
% \label{tab:DecFor}
% \end{table}
% \textit{Abbreviations: PBC = Patient-based contrastive, ABC = Anomaly-based contrastive, ARIMA = Autoregressive integrated moving average  }

\begin{table*}[t]
\centering
\caption{Overview of the performance of predicting timelines leading to hospitalization for the tested architectural setups trained on event-based detection in the study. \\ \textit{Abbreviations: PBC = Patient-based contrastive, ABC = Anomaly-based contrastive, ARIMA = Autoregressive integrated moving average.}}
\begin{tabular}{p{4cm}p{2.5cm}p{1.3cm}p{1.3cm}p{1.3cm}p{1.3cm}}
\hline\hline
Model type (Pre-training) & Parameter count & Sensitivity & Specificity & FA-PPM & NNE \\
\hline
Transformer (No pre-train) & 25837412 & 0.629 & 0.910 & 2.86 & 24.87 \\
TRACER & 25837412 & 0.667 & 0.921 & 2.49 & 20.79 \\
Transformer (PBC) & 25837412 & 0.037 & 0.937 & 2.01 & 286.53 \\
Transformer (PBC\&ABC) & 25837412 & 0.518 & 0.909 & 2.86 & 30.14 \\
Transformer (ABC\&PBC) & 25837412 & 0.370 & 0.935 & 2.04 & 30.00 \\
LSTM & 253635 & 0.480 & 0.932 & 2.14 & 16.12 \\
T-LSTM & 17461599 & 0.520 & 0.926 & 2.33 & 15.87 \\
ARIMA/XGBoost & 48114 nodes&  0.444 & 0.902 & 3.08 & 37.74 \\
\hline\hline
\end{tabular}
\label{tab:DecFor}
\end{table*}

\subsubsection{Training with timeline prediction}
Continuing, we evaluated the performance of different model architectures when both training and validation were performed on predicting timelines leading to hospitalization (Table \ref{tab:ForFor}). Overall, all evaluated architectures demonstrated reduced sensitivity compared to the models that were trained using the event-based detection formulation, (Table \ref{tab:DecFor}), indicating a difficulty for the models to reliably predict timelines for hospitalization without the additional views of the events. Compared to the event-based training, the TRACER model trained on timeline prediction had a lower overestimation of hospitalization using timeline prediction training which led to lower FA-PPM and NNE scores. Notably, this conservative prediction behavior also resulted in a 61.1\% reduction in sensitivity for the model. The ABC\&PBC pre-trained transformer failed to correctly predict a timeline leading to hospitalization in the validation data, resulting in no NNE score despite achieving high specificity.

\begin{table}[H]
\centering
\caption{Overview of the performance of predicting timelines leading to hospitalization for the tested architectural setups trained on timeline prediction in the study. \\ \textit{Abbreviations: PBC = Patient-based contrastive, ABC = Anomaly-based contrastive, ARIMA = Autoregressive integrated moving average  }}
\begin{tabular}{p{3.1cm}p{0.9cm}p{0.9cm}p{1cm}p{0.9cm}}
\hline\hline
Model type (Pre-training) & Sensitivity & Specificity & FA-PPM & NNE \\
\hline
Transformer (No pre-train)  & 0.222 & 0.905 & 2.99 & 71.94   \\
TRACER & 0.259 & 0.976 & 0.76 & 16.58\\
Transformer (PBC)  & 0.037 & 0.978 & 0.69 & 98.03 \\
Transformer (PBC\&ABC)  & 0.111 & 0.945 & 1.73 & 83.33 \\
Transformer (ABC\&PBC)  & 0 & 0.985 & 0.47 & - \\
LSTM  & 0.149 & 0.945 & 1.73 & 79.44 \\
T-LSTM  & 0.071 & 0.941 & 1.86 & 227.27 \\
ARIMA/XGBoost  & 0.074 & 0.906 & 2.96 & 212.76 \\
 \hline\hline
\end{tabular}
\label{tab:ForFor}
\end{table}

\subsection{Ablation studies}

\subsubsection{Performance with added weight measurements}
\noindent We studied the individual influence on performance of the included biomarkers for predicting timelines leading to hospitalization. For this, we trained the TRACER model with one biomarker at a time. Moreover, we also included the recorded weights of the patients as a fourth biomarker. Following this, we compared the performance of the setup in our main findings using HR, SYS and DIA measurements to that of a system including weights as well (Table \ref{tab:VPs}). We observed similar performance for the models trained solely on variables reflecting cardiovascular physiology. The model trained on only weight was not able to detect the hospitalization event on par with the other variables. Moreover, training the TRACER model including the weight information resulted in substantially reduced performance on correctly identifying timelines leading to hospitalization and a higher NNE, with a minor decrease in overestimation to the original TRACER model.

\begin{table}
\centering
\caption{Overview of the performance of predicting timelines leading to hospitalization for each biomarker and with added weight measurements in the study with TRACER. \\ \textit{Abbreviations: HR = Heart Rate, SYS = Systolic Blood Pressure, DIA = Diastolic Blood Pressure, W = Weight}}
\begin{tabular}{p{1.3cm}p{1.5cm}p{0.9cm}p{0.9cm}p{1cm}p{0.8cm}} \hline\hline
&Biomarker & Sensitivity &Specificity  & FA-PPM & NNE \\
\hline
One at  & HR & 0.667 & 0.924 & 2.39 & 20.16 \\
a time &SYS  & 0.667 & 0.917 & 2.61 & 21.69 \\
&DIA  & 0.667 & 0.921 & 2.48 & 20.70 \\
&W & 0.111 & 0.961 & 1.23 & 56.18 \\
\hline
&Model &  &   &   \\
\hline
Combination &TRACER + W & 0.389 & 0.927 & 2.29 & 30.39  \\
&TRACER & 0.667 & 0.921 & 2.49 & 20.79  \\
 \hline\hline
\end{tabular}
\label{tab:VPs}
\end{table}

\subsubsection{Patient-based hospitalization detection}
\noindent We evaluated the TRACER model and a baseline transformer model without pre-training on how well each model predicts timelines leading to hospitalization per biomarker and hospitalized patient in the validation set (n=4 patients). One patient did not record sufficient measurement information to form a window before the hospitalization event, leaving the remaining three patients available for analysis (Table \ref{tab:patients}). Overall, Patients 2- and 3 were detected by both models. The ABC pre-training in TRACER increased the sensitivity of Patient 2 for the DIA biomarker leading to the observed increase in overall sensitivity (Table \ref{tab:DecFor}). However, Patient 1 went undetected by both models for all biomarkers.

\begin{table}
\centering
\caption{Overview of the biomarker-specific sensitivity of predicting timelines leading to hospitalization for each hospitalized patient in the validation set. \\ \textit{Abbreviations: HR = heart Rate, SYS = systolic blood pressure, DIA = diastolic blood pressure  }
 }
\begin{tabular}{l|ccc|ccc} 
    \hline\hline
    Patient \# & \multicolumn{3}{c|}{Sensitivity (TRACER)} & \multicolumn{3}{c}{Sensitivity (Transformer)} \\
    \hline
    Biomarker & HR & SYS & DIA & HR & SYS & DIA \\
    \hline
    Patient 1 & 0 & 0 & 0 & 0 & 0 & 0 \\
    Patient 2 & 0.833 & 0.833 & 0.833 & 0.833 & 0.833 & 0.667 \\
    Patient 3  & 1 & 1 & 1 &  1 & 1 & 1 \\
    \hline\hline
\end{tabular}
\label{tab:patients}
\end{table}

\subsubsection{Performance with varying encoder complexity}
\noindent We studied the performance of different complexity levels in the encoders for the TRACER model by varying the number of attention blocks in each encoder (Table \ref{tab:complexity}). We observed the lowest performance when only one attention block was used in the encoder compared to two- or more attention blocks. The use of three or more attention blocks was similar to one another, with four attention blocks performing marginally better at  detecting timelines leading to hospitalization.

\begin{table}
\centering
\caption{Overview of the performance of predicting timelines leading to hospitalization for different numbers of attention blocks in the transformer encoders. }
\begin{tabular}{p{1cm}p{1cm}p{1cm}p{1cm}p{1cm}p{1cm}}
\hline\hline
\# Blocks & Parameters & Sensitivity & Specificity & FA-PPM & NNE \\
\hline
1 & 8976740 & 0.185 & 0.886 & 3.59 & 103.30   \\
2 &17407076 & 0.481 & 0.924 & 2.39 & 26.88 \\
3 & 25837412& 0.667 & 0.921 & 2.49 & 20.79 \\
4 & 34267748& 0.704 & 0.919 & 2.55 & 20.20 \\
5 & 42698084& 0.667 & 0.918 & 2.58 & 21.46 \\

 \hline\hline
\end{tabular}
\label{tab:complexity}
\end{table}

\subsubsection{Performance across varying seeds}
We assessed the robustness of model performance to variations in the selected hospitalized patients in the validation set by generating five randomized seeds that affected both training randomization and patient shuffling in the training and validation sets. We reported the average performance across five randomized seeds for the TRACER model and a baseline transformer without pre-training (Table \ref{tab:seeds}). Similar average specificity was observed for TRACER and the baseline transformer. TRACER achieved a higher average sensitivity across the five seeds compared with the baseline transformer, indicating better average prediction ability of hospitalization events. The NNE varied substantially across seeds, resulting in a high standard deviation for both models. FA-PPM also showed variability across seeds, although the average values remained similar between the two models.

\begin{table}[H]
\centering
\caption{Overview of the average performance for five different seeds. \\ 
\textit{Abbreviations: std = standard deviation}}
\begin{tabular}{p{2.4cm}p{2cm}p{3.2cm}}
\hline\hline
Metric (mean $\pm$ std)  & TRACER & Transformer (No pre-train) \\
\hline
Sensitivity & 0.537 $\pm$ 0.15 & 0.441 $\pm$ 0.13 \\
Specificity & 0.921 $\pm$ 0.04 & 0.931 $\pm$ 0.03 \\
FA-PPM & 2.50 $\pm$ 1.12 & 2.15 $\pm$ 1.07 \\
NNE & 101.10 $\pm$ 67.66 & 123.09 $\pm$ 110.20  \\
\hline\hline
\end{tabular}
\label{tab:seeds}
\end{table}

\section{Discussion}
\noindent The results from the study show that the transformer-based models performed better in predicting timelines leading to worsening HF, as represented by the need of hospitalization, in the low-resolution irregularly sampled telemonitoring data compared to other tested ML and DL models. Primarily, the TRACER model (Transformer with ABC pre-training) showed superior performance compared to the other transformer models and the LSTMs (Table \ref{tab:DecFor}). Training TRACER on an event-based detection task rather than forecasting hospitalization events improved its ability to identify timelines leading to hospitalization, suggesting that utilizing multiple temporal contexts around the limited hospitalization events enables more effective learning from sparse telemonitoring data.

The TRACER model is a large model built with a parameter count in the region of $10^{7}$. This affects its usability in training- and runtime compared to a more shallow network such as the LSTM model with roughly 250 000 parameters. However, the results show that model performance is influenced by both encoder complexity and the applied training strategy (Table \ref{tab:DecFor} and Table \ref{tab:complexity}). Increasing the number of encoder blocks improved performance up to three blocks, after which additional blocks in the transformer encoder provided limited improvements. Notably, the results also indicate that there is a complexity threshold requirement for the transformer encoders, as a TRACER model with only one encoder block performed worse than the ARIMA/XGBoost baseline. 

% The model with the lowest performance in detecting hospitalization events was the ARIMA/XGBoost which was tailored to a more simple solution. The addition of ARIMA parameters to the window measurement values attempted to maintain some temporal properties of the input. However, ARIMA assumes evenly spaced observations in the time series which could have an association with the lower performance \cite{schaffer2021interrupted}.

 \indent The four routinely collected physiological variables contributed differently to model performance. Weight showed the poorest performance when modelled alone and reduced overall performance when added to the three cardiovascular variables (Table \ref{tab:VPs}). Notably, heart rate and blood pressure contributed more strongly to prediction by our model, whereas international HF guidelines specifically recommend monitoring unexpected weight gain, with an increase of $>$2 kg within 3 days prompting diuretic adjustment and/or contact with the HF team \cite{ESC2021}. However, weight measurements were more inconsistently and irregularly recorded (Figure \ref{fig:physiometric}), and these findings should therefore not be interpreted as evidence that weight lacks clinical value. The comparatively limited predictive value of weight in our cohort may reflect measurement quality, the small number of events, and a population largely comprising patients with newly diagnosed HF undergoing optimization of disease-modifying therapy rather than treatment of overt congestion. These findings support the need for development of multivariable prediction models using temporal patterns in routinely collected measurements during contemporary HF follow-up. \\
 \indent The sparse time resolution of the intervals from the daily recordings, irregular sampling of data and the low proportion of outcomes may limit the performance and possible tasks. An imbalance ratio of 1244.63 in the dataset makes any detection task difficult as it is considered to be an extremely rare event \cite{rareevent}. Although TRACER achieved the highest sensitivity among the evaluated models, it still generated approximately 2.5 false alerts per patient-month and required about 21 clinical evaluations to identify one true hospitalization. Given the low incidence of hospitalization events, this alert burden could cause cognitive overload, unnecessary assessments, and alert fatigue, potentially reducing clinicians responsiveness to meaningful warnings \cite{ancker2017effects}. However, this should be weighed against current practice, in which an HF nurse reviews every single observation from daily measurement and acts based on clinical judgment.\\
\indent The proposed modeling approach should be interpreted as a clinical decision-support tool for risk stratification, identifying measurement patterns associated with upcoming HF-related hospitalization rather than a diagnostic system for detecting acute deterioration. Since hospitalization events may arise from heterogeneous clinical trajectories, the identified patterns may reflect a combination of physiological changes preceding hospitalization and other temporal changes occurring before the event.  The potential clinical role of the tool depends on identifying the HF population in which it is most useful. During the six-month monitoring period, relatively few patients were hospitalized: 56 overall (20.3\%) and 18 for HF-related causes (6.5\%), whereas higher hospitalization rates would be expected in a general Swedish HF population \cite{Boman2021Healthcare}. This imbalance may reflect a comparatively stable cohort, including a majority of patients undergoing first-time initiation or optimization of guideline-directed medical therapy and thus likely to improve during follow-up. For this type of HF population, measurement-based telemonitoring provides an accessible approach for longitudinal monitoring without requiring additional procedures or dedicated implantable infrastructure, emphasizing scalability and broader patient coverage over the higher-resolution physiological data available from implantable monitoring systems. In such a population, a model’s greatest value may lie in reducing unnecessary clinical workload, particularly if reporting leverages its high specificity to identify patients unlikely to require medical action. This potential benefit may add value compared with current routine follow-up, in which HF nurses review each individual registered measurement.\\
\indent The two pre-training methods had different impacts on model performance. ABC pre-training enhanced the model’s ability to predict timelines leading to hospitalizations, whereas introducing PBC pre-training generally reduced it. When PBC was applied before ABC, the ABC training could refine the embeddings, yielding comparable sensitivity to the LSTM models with reduced specificity. However, reversing the order caused PBC to overwrite the ABC embeddings, leading to worse sensitivity than the LSTM models. Possible reasons to explain why PBC pre-training limits the system include the small number of subjects and few hospitalization events, which may confuse the model when hospitalized patients closely resemble non-hospitalized ones. Additionally, using all 58 variables to create patient profiles combined with a 99\% similarity threshold may be too lenient for forming positive pairs. Since many profile segments contained missing values replaced with the same padding, dissimilar patient profiles could appear artificially similar, affecting the models learning. \\
\indent Although the cohort included patients from three hospitals in Region Västra Götaland, the generalizability of the model to other healthcare settings, populations and measurement devices remains uncertain. Differences in patient case-mix, clinical workflows, measurement routines, device characteristics and data completeness may affect model performance. External validation in an independent dataset, together with local calibration and prospective clinical evaluation, will therefore be required before broader implementation.

\section{Conclusion}
\noindent In conclusion, the TRACER model demonstrated superior performance in predicting timelines leading to worsening HF in our study of low-resolution telemonitoring data compared to the other ML- and DL-based implementations. Applying pre-training techniques was observed to both boost- and harm the performance of the system. Training TRACER on a reformulated event detection task can enable more effective use of the limited hospitalization events available in our imbalanced real-world dataset.

\section*{References}

\bibliographystyle{IEEEtran}
\bibliography{referenced}

\end{document}